# Display of Information for Time-Critical Decision Making

Eric Horvitz
Decision Theory Group
Microsoft Research
Redmond, Washington 98025
horvitz@microsoft.com

Matthew Barry*
Propulsion Systems Section
NASA Johnson Space Center
Houston, Texas 77058
barry@rpal.rockwell.com

## Abstract

We describe methods for managing the complexity of information displayed to people responsible for making high-stakes, time-critical decisions. The techniques provide tools for real-time control of the configuration and quantity of information displayed to a user, and a methodology for designing flexible human-computer interfaces for monitoring applications. After defining a prototypical set of display decision problems, we introduce the expected value of revealed information (EVRI) and the related measure of expected value of displayed information (EVDI). We describe how these measures can be used to enhance computer displays used for monitoring complex systems. We motivate the presentation by discussing our efforts to employ decision-theoretic control of displays for a time-critical monitoring application at the NASA Mission Control Center in Houston.

## 1 INTRODUCTION

The rapid growth in the use of computers to access information and to monitor complex systems has brought increased attention to the costs of navigating through large quantities of data in search of critical information. Problems with accessing and reviewing information are especially salient in high-stakes, time-critical decision-making contexts. We present work on enhancing the human–computer interface through applying decision-theoretic inference to control the information displayed to people responsible for monitoring complex systems. The methods can be employed to enhance the quality and timeliness of decisions by adjusting the configuration and quantity of information displayed to decision makers, depending on the current uncertainties and time criticality. Beyond their application in the real-time control displays, the techniques can also assist engineers with the offline design of user-interface content and functionality. We will see also how a decision-theoretic perspective on display highlights important questions about human information processing. Answers to these questions will allow us to further refine the display-management methodology.

Previous related investigation of the use of probability and utility in display management includes work on controlling the tradeoff between the completeness and the complexity of informational displays, computer-based explanations, and computational behavior [Horvitz et al., 1989], the use of multiattribute utility to control the complexity of presentations and displays [Horvitz, 1987a, Mclaughlin, 1987], and multiattribute utility for queuing and prioritizing the results of diagnostic reasoning [Breese et al., 1991]. In other related work, qualitative models, in combination with several heuristic importance metrics have been employed to select information for monitoring applications [Doyle et al., 1989].

We will first review basic results from studies of human information processing about cognitive load, short-term memory, and decision making. Then, we will describe the representation and solution of time-critical decision problems. We will introduce the task of monitoring and decision making about propulsion systems on the Space Shuttle to motivate the importance of information display in time-critical situations. We will present decision models for the display of information, and discuss methods for evaluating the value of displayed information. We will focus first on the expected value of revealed information (EVRI). After, we will introduce the use of Bayesian models of user belief and action, and describe the expected value of displayed information (EVDI). Finally, we will address practical approaches to implementing display managers based on EVRI and EVDI. We will summarize by discussing our research directions.

## 2 INFORMATION AND COGNITIVE LIMITATIONS

* Current address: Rockwell Space Operations Company, Mail Code R20A-4, 600 Gemini, Houston, TX 77058.

Why should we worry about managing the complexity of displayed information? Decision-theoretic analyses of the value of information show us that gaining cost-free access to additional information can only enhance the quality of our actions. Unfortunately, information often comes at a cost. When we compute the net value of information, we consider the value of information, given uncertainty about test results, and the cost of information.

Information that is already available to a computer about a monitored system typically does not cost anything to display. However, in time-critical, high-stakes situations, the time required by people to review information, and confusion arising in attempts to process large amounts of data quickly, can lead to costly delays and errors.

Fundamental limitations in the abilities of people to process information explain why decision quality may degrade with increases in the quantity and complexity of data being reviewed, and with diminishment in the time available for a response. Human difficulties with the processing of information has been a key research focus within Cognitive Psychology [Bruner et al., 1956]. Numerous studies have provided evidence that human information processing is primarily sequential in nature [Simon, 1972]. Experiments have shown that the speed at which subjects perform tasks drops as the quantity of information being considered increases, and that the rate of performing tasks can be increased by filtering or suppressing irrelevant information [Morrin et al., 1961]. In a classic study on limitations in human cognition, Miller found that humans cannot consider more than five to nine distinct concepts or "chunks" of information simultaneously [Miller, 1956]. The capacity of decision makers to consider important influences on a decision may be reduced even further if fast action is demanded in crisis situations. One cognitive-psychology study demonstrated that people cannot retain and reason simultaneously about more than two concepts in environments filled with distractions [Waugh and Norman, 1965]. These and other cognitive psychology findings provide motivation for automated methods that can balance the value and costs of displayed information.

## 3 TIME-CRITICAL DECISIONS

The costs versus the benefits of spending time to review additional information are sensitive to the time criticality of a situation. In time-critical contexts, the utilities of outcomes diminish significantly with delays in taking appropriate action. There are several classes of time-dependent decision problems and a variety of ways to represent variables and probabilistic dependencies to encode knowledge about time-dependence of outcome and utility [Horvitz, 1987b, Horvitz, 1988]. In one approach, we model explicitly the time-dependent progression of important states of a system under the influence of processes that may persist over time. We consider the effects of different actions (including not taking any explicit action) at different times on the temporal progression of the states of a system. In some cases, it may be appropriate to assess an outcome as an equilibrium state, reached at some time following a set of interventions, and to assess the utility of this equilibrium state [Horvitz, 1990]. We can attempt to minimize the detailed modeling of the temporal progression of system states by assessing the time-dependent changes in the utility of outcomes that are defined in terms of a system (or world) state and interventions made at various times. The utility associated with allowing one or more states of a system to evolve with or without intervention is a function of the state, the action, and the time action is taken. It may be possible to assess from experts time-dependent utility functions that capture the changes in utility at progressively later times of intervention for different system states [Horvitz and Rutledge, 1991].

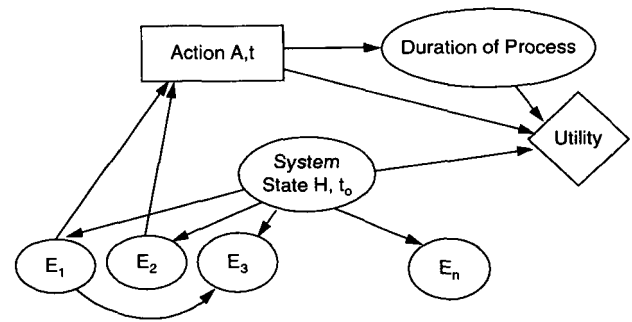

Figure 1: An influence diagram representing the time-dependent cost associated with allowing an anomalous condition to persist by delaying an effective response.

In one class of time-critical decision problem, representing a large class of high-stakes, time-critical monitoring tasks, we gain access to information about system's behavior at time $t_0$ and attempt to diagnose and take action to respond to the anomalous state of the system. In many time-critical situations, additional information about the progression of a state over time is either not relevant or not available before action needs to be taken. Outcomes are often particularly sensitive to the length of time that an anomalous condition persists. Delays may be associated with significant time-dependent changes in the utility of outcomes based in the duration of the condition. A representation of this type of decision problem, assuming a single action or fixed sequence of actions, is represented by the influence diagram in Figure 1. We assess time-dependent utilities, as a function of the action $A_i$, the state of the system $H_j$, and the delay before action is taken, $u(A_i, H_j, t)$. Given uncertainty about the state of system, the expected utility (EU) of taking action $A_i$ at time $t$ is

$$\text{EU}(A_i, t) = \sum_{j=1}^{n} p(H_j|E, \xi) u(A_i, H_j, t)$$

where $p(H_j|E, \xi)$ is the probability over hypotheses

298    Horvitz and Barry

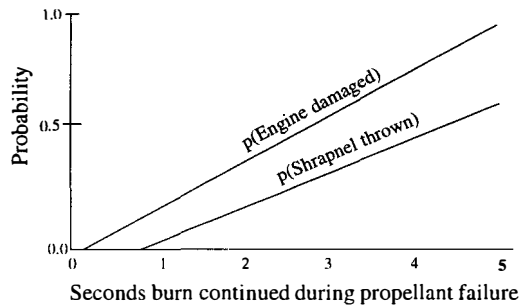

Figure 2: Graphs of the probabilities of costly failures with delayed action, highlighting the potential time criticality of decision making about Shuttle propulsion systems.

about different systems states, given observations $E$ and background state of information $\xi$.

## 4  EXAMPLE: SPACE SHUTTLE PROPULSION SYSTEMS

Our work on display management for time-critical decisions has been a key component of the multi-site Vista Project [Horvitz et al., 1992]. The Vista Project was initiated in 1991 to develop inferential tools to assist flight engineers at the NASA Mission Control Center in Houston with the interpretation of telemetry from the Space Shuttle. We designed and implemented decision-theoretic inference and display-management software to aid engineers monitoring the Shuttle's propulsion systems. To further motivate display-management issues, we will review some detail about the propulsion-systems monitoring and decision problem.

Flight engineers in the Propulsion Section at Johnson Space Center are responsible for monitoring two different Shuttle thruster systems: the *orbital maneuvering system* (OMS) and the *reaction control system* (RCS). The large right and left OMS engines are fired for such critical maneuvers as orbital insertion and orbit circularization. The smaller suites of RCS thrusters are used for translation in space, for such tasks as maneuvering near objects in orbit, as well as for the continual computer-controlled stabilization of the Shuttle's trajectory and position.

Flight engineers often face a large quantity of potentially relevant information, especially during crises. Propulsion flight engineers must continue to monitor multiple sensors which measure such variables as changes in the Shuttle's velocity with burns, pressures and temperatures in tanks of consumables (helium, fuel, nitrogen, and oxidizer), and voltages and currents in electrical subsystems.

If a problem with the functioning of the propulsion systems is noted during a critical burn, the operator must decide whether to continue the burn, halt the burn, or redirect fuel to alternative engines in a variety of different ways. The stakes may be high. For example, continuing a burn during an engine problem can destroy the engines or the entire Space Shuttle. Halting a burn before a critical target velocity is reached can lead to such situations as the forced ditching of the orbiter in the ocean, or the missing of a critical key reentry opportunity. Decisions are not only high-stakes, they may also be time critical. In many contexts, delayed decisions can be very costly to the mission and to the orbitor itself. The time criticality of propulsion-systems decision making is highlighted by the graph in Figure 2, displaying dynamic probabilities, assessed from an expert, of damaging and destroying an engine explosively by continuing a burn during a propellant failure.

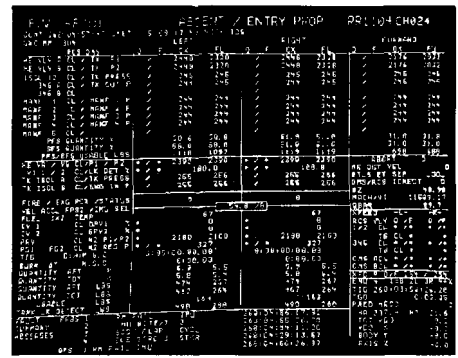

Figure 3: Primary screen of the traditional display for Shuttle propulsion systems at Mission Control. Left, right, and forward RCS data is displayed in panels at the top of the screen. Data on the right and left OMS engines are displayed in the lower panels.

### 4.1  STATUS QUO FOR DISPLAY

Before Vista, the traditional computer displays for the Propulsion Section resembled other cluttered, information-rich displays at the Mission Control Center. The primary propulsion-systems display, in use before the Vista system was introduced, is pictured in Figure 3. The screen includes information on the status of two OMS engines, and the three banks of RCS engines. During missions, if ground controllers become concerned about the health of one of the propulsion systems or subsystems, auxiliary screens may be requested which contain such data as trend information about engine consumables. The complex primary and auxiliary displays of information can become burdensome in situations that demand quick decision making, especially for flight engineers in training.

### 4.2  DECISION MODELS FOR PROPULSION SYSTEMS

In the first phase of the Vista project, we developed probabilistic and decision-theoretic models for assist-ing flight engineers with key subproblems in the Shuttle propulsion domain. First, we developed Bayesian networks for the Shuttle's propulsion systems. Figure 4 displays a basic schematic of the OMS engine. The overall operation of the propulsion engines is straightforward. A tank of helium gas maintains pressure on tanks of oxidizer and fuel. To fire an OMS engine, valves are opened which allow the fuel and oxidizer to mix and combust to provide thrust. In addition to the basic flows, ground controllers must also consider the status of a set of valves between various tanks, and crossover lines that allow propellant to be shared by different engine systems. Suites of temperature and pressure sensors are located at critical locations in the system. The telemetry about propulsion systems transmitted to ground stations consists largely of information from these sensors.

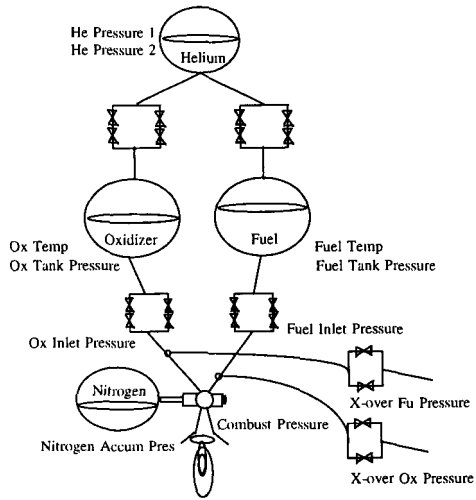

Figure 4: Schematic of an OMS engine.

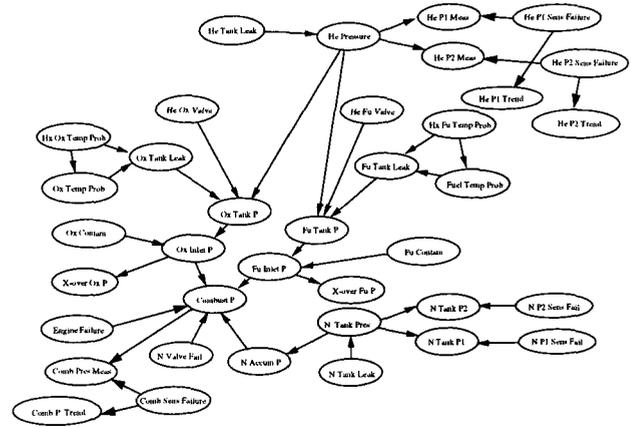

Figure 5: A Bayesian network for the OMS engine.

Figure 5 shows the graphical structure of a Bayesian network for an OMS engine. The Bayesian networks that we constructed for Shuttle propulsion systems are notable in that they include rich representations of sensor failures and errors. The models consider failures of sensors as well as failures of core components of propulsion systems. The sensor-error models included in the Bayesian networks represent information about the validity of sensors as well as the way that different sensors fail. Experts with long-term experience make use of evidence about sensor failure, such as noting that a specific class of sensor is generating a sinusoidal output or is trending upward when a physical model only makes possible decreasing quantities of the measured substance.

### 4.3 REPRESENTING ACTION AND TIME

Moving from inference about anomalous states to the realm of actions, we modeled decisions and outcomes for different anomalies and contexts. We enumerated available actions and considered the time-dependence of outcomes. We assessed time-dependent utility in terms of the dynamically changing probability that a mission would be terminated prematurely or, for more catastrophic situations, that the entire orbitor would be lost, as a function of the anomaly, the action taken, and the persistence of fault states. We found that it was useful to use multiattribute utility to understand the tradeoffs among dimensions of value in an outcome, including such key attributes as the portion of the target velocity reached and the probability of damaging an engine being used for the desired impulse.

The Vista-II system went into service at the Mission Control Center in 1993, following laboratory validation with a prototype named Vista-I. Vista-II monitors and interprets live telemetry being transmitted from the Shuttle. When the probability of an anomaly (including sensor faults) exceeds a small threshold, the system displays a list of possible faults ranked by likelihood with an associated graphical display of the probabilities of the faults. In addition to providing a probability distribution over faults, the system also generates recommendations about ideal action. A list of possible actions, ranked by expected utility is displayed.

## 5 DISPLAY DECISION MAKING

Several approaches to the management of display and information access were implemented in the initial Vista systems [Horvitz et al., 1992]. These include means for flexibly controlling the detail of information presented about specific subsystems depending on the context and inference about anomalies, the use of a list of faults, sorted by probability, as an active index into related trend information, and the prioritization of faults by the expected cost of delay to review the faults.

To build a flexible approach to display, data templates—contiguous sets of related information— were designed for each subsystem (the two OMS systems and the three RCS systems). A spectrum of tem-plates were created for each subsystem, spanning a range of completeness from the most detailed to progressively less complete, more abstract presentations of data. We also introduced levels of detail in the display of diagnostic information, by allowing the system to display probabilities of anomalies at more abstract levels of the subsystems, such as the probability of a sensor failure versus a specific sensor failing.

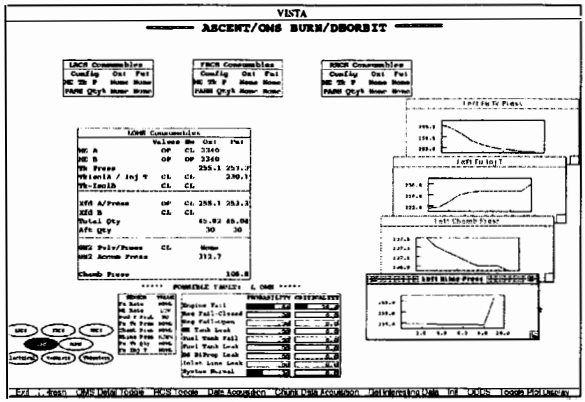

Figure 6: A screen from Vista-I displaying how a problem with the left OMS is handled. The left-OMS data template has been expanded from the core data set by the display manager. Relevant trend information is displayed on the right side of the screen.

In the main Vista display, templates for the different subsystems are configured in an invariant pattern, introducing spatial stability in the location of information about different subsystems. At run time, the results of inference are used to modify the amount of detail displayed about each subsystem; the data template for a subsystem appears to telescope from a compact summary into a larger, more complete presentation. The overall positioning of the data templates for different subsystems remains the same during the resizing of presentations on specific subsystems, minimizing the effort needed to locate information. Preferred levels of detail for each subsystem were predefined for different contexts (e.g., critical OMS burn, orbital coast, etc.). Beyond automated control, we designed the interface to allow users easy manual access to any information available to the Vista system.

The combination of decision-theoretic inference with flexible access to a range of detail on subsystems has worked well in Vista. Nevertheless, we have continued to pursue principles for controlling displays. We will now focus on newer methods, several of which are being validated for the forthcoming Vista-III system for the Mission Control Center.

## 5.1   EXPECTED VALUE OF REVEALED INFORMATION

The goal of display management in time-critical situations is to maximize the expected utility of an operator's decisions. We now explore methods to characterize the costs and benefits of displaying different configurations of information.

Let us first consider a quantity we refer to as the expected value of revealed information (EVRI). EVRI is the expected value of considering additional quantities of information that is available with certainty, yet is hidden from a decision analysis. The expected utility of considering previously hidden information must be evaluated in the context of a complete decision-theoretic analysis, taking advantage of *all* of the available information. EVRI differs from the expected value of information (EVI) in that EVI includes a consideration of uncertainty about the state of observations. In the case of EVRI, an automated display manager has access to all available data. EVI is appropriate metric for display in cases where a system must expend effort before access is gained to sensor values.

Consider the case of monitoring complex systems such as Shuttle propulsion systems. Human operators are charged with reviewing data that is typically accessed through a battery of sensors that innervate a monitored system. We will consider use of a display-management system to make decisions about the nature and quantity of evidence $E$ to be displayed. However, we could apply a similar analysis for controlling the display of information about other distinctions represented in or inferred from a decision model.

Assume that a monitoring system has access to a set of sensed observations, $\mathbf{E}$. The probabilities over hypotheses of interest $H$ (*e.g.*, failures in a monitored system), inferred with a gold-standard diagnostic model that takes into consideration all available data, $p(H|\mathbf{E},\xi)$, can be used to compute the expected utility of the gold-standard action, $A^{G*}$,

$$A^{G*} = \arg\max_A \sum_j u(A_i, H_j) p(H_j|\mathbf{E},\xi) \quad (1)$$

Let us now *hide* some evidence from the analysis and consider, with the same decision model, the value of revealing or *displaying* a subset of observations, $E \subset \mathbf{E}$. We compute a potentially revised optimal action, $A^{D*}$, based on the revised probability distribution, $p(H|E,\xi)$. We compute the best action by substituting the revised probability distribution into Equation 1,

$$A^{D*}(E) = \arg\max_A \sum_j u(A_i, H_j) p(H_j|E,\xi) \quad (2)$$

We must evaluate the expected utility of $A^{D*}$ with the gold-standard probability distribution, considering *all* of the available evidence $\mathbf{E}$,

$$\mathrm{eu}[A^{D*}(E)] = \sum_j u[A^{D*}(E), H_j] p(H_j|\mathbf{E},\xi) \quad (3)$$

We can now define the expected value of revealed information. The $\mathrm{EVRI}(e, E, \mathbf{E})$ is the expected value of revealing a set of additional information $e$ in a context defined by the set of previously revealed information $E$ and the set of all available evidence $\mathbf{E}$. Note that the EVRI is zero if the action does not change with the revealed information.

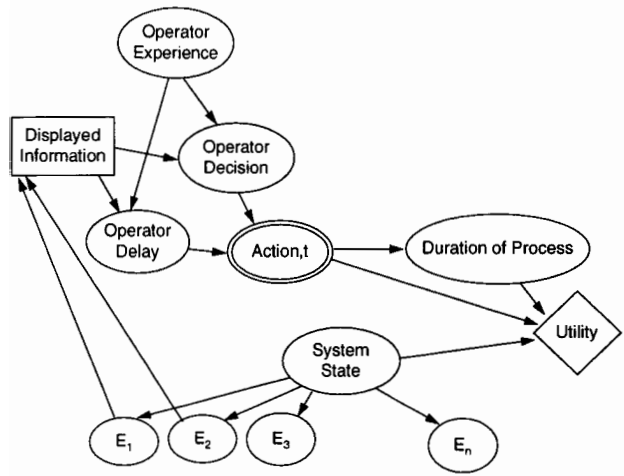

Figure 7: An influence diagram representing the problem of controlling the display of information in time-critical situations. We represent user decisions as chance variables.

$$\text{EVRI}(e, E, \mathbf{E}) = \text{eu}[A^{D*}(E+e, \mathbf{E})] - eu[A^{D*}(E, \mathbf{E})] \quad (4)$$

EVRI does not take into account the costs potentially associated with the review of increasing quantities of information. Action may be delayed if a decision-making agent must process additional information. Such time delays may change the best action or incur significant losses in the maximum expected utility in time-critical settings. The *net* expected value of revealing information (NEVRI) includes the costs and benefits of reviewing the additional information. Let us assume that costs are based solely in deterministic delays $t(e)$ required to review information $e$. The best decision given consideration of only evidence $E$ is,

$$A^{D*}(E) = \arg\max_A \sum_j u[A_i, H_j, t(E)] p(H_j|E, \xi) \quad (5)$$

The expected value of this decision is,

$$\text{eu}[A^{D*}(E)] = \sum_j \text{eu}[A^{D*}(E), H_j, t(E)] p(H_j|\mathbf{E}, \xi) \quad (6)$$

conditioning the probability of states of the system on the complete set of available evidence. The NEVRI can be computed by considering the best actions $A^{D*}$ and expected utilities of these actions, given a consideration of $t(E+e)$ versus $t(E)$.

$$\text{NEVRI}(e, E, \mathbf{E}) = \text{eu}[A^{D*}(E+e)] - \text{eu}[A^{D*}(E)] \quad (7)$$

Note that we can easily generalize NEVRI to include information about the uncertainty associated with the time required to review information.

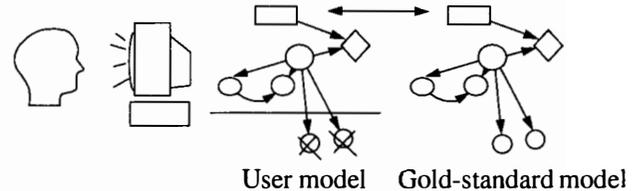

Figure 8: We employ a user and a gold-standard decision model to determine the value of displaying additional information. The user model can be a gold-standard model acting on a subset of instantiations or a model representing a user's causal knowledge and preferences.

Let us consider the situation where a subset of all information (*e.g.*, a subset of relevant telemetry from the Shuttle) is revealed by an automated display manager to a decision maker charged with responsibility for making a decision. For now, we assume that the decision maker is an expert who acts in accordance with a gold-standard diagnostic model but requires increasing amounts of time to review larger quantities of information. We can use NEVRI to consider the costs versus benefits of displaying alternate subsets of available information. We can search through all configurations of evidence to find a subset of information, $e^*$, that maximizes the expected utility,

$$e^* = \arg\max_e \text{NEVRI}(e, E, \mathbf{E}) \quad (8)$$

For the general case, finding the best subset of monitored information to display requires a search over all combinations of data. We will discuss practical strategies that coincide with extensions to current Vista display policies in Section 7. First, we will generalize EVRI to the expected value of displayed information (EVDI).

### 5.2 GENERALIZATION TO VALUE OF DISPLAYED INFORMATION

EVRI considers the costs and benefits of revealing subsets of all available observations in the context of a gold-standard model. In the general case, we cannot assume that a user will act in accordance with a gold-standard diagnostic or decision model. A novice decision maker may make suboptimal decisions relative to a gold-standard decision-theoretic model yet still be relied upon for action given traditional desires for having a human in the loop. Alternately, we may wish to use automated display management to train a user about the most important information to consider. We now generalize the EVRI to the expected value of displayed information (EVDI) by considering an operator's actions in response to varying quantities of displayed information and the value of this information to the user from the perspective of the gold-standard model.

We wish to automate decisions about display to optimize the expected value of an operator's actions. Al-though we can consider several different forms of information, including the output of automated inference, we shall again cast the discussion in terms of decisions about the display of subsets of all monitored observations, $E \subseteq \mathbf{E}$. To make a decision about the nature and quantity of evidence to reveal to the system user, we consider the likelihood of alternative user actions and likelihoods of delay as functions of displayed data.

Within the context of the time-critical decision problems broadly captured by the influence diagram in Figure 2, the problem of displaying information in time-critical settings is represented by the influence diagram displayed in Figure 7. We transform the human operator's decision and delay to chance variables that are influenced by the information displayed. We represent the user's expertise or background by conditioning the action and delay nodes on a variable representing expertise. If we have certain knowledge about the user's background, we condition the decision node on this information, represented by an information arc.

To simplify our equations, we will assume that the delay in the human operator's action is a deterministic function of the quantity of evidence displayed and is independent of the ultimate action taken. We will keep the operator's expertise implicit. Given an initial display of system observations $E$, and a complete set of evidence $\mathbf{E}$ known to a display manager, the net value of displaying additional information (EVDI) $e$ is,

$$\mathrm{EVDI}(e, E, \mathbf{E}) = \\ \sum_i p(A_i|E, e, \xi) \sum_j u[A_i, H_j, t(E+e)] p(H_j|\mathbf{E}, \xi) \\ - \sum_i p(A_i|E, \xi) \sum_j u[A_i, H_j, t(E)] p(H_j|\mathbf{E}, \xi) \quad (9)$$

We can generalize EVDI to include a consideration of other factors including uncertainty about relevant information that a user may already know or has acquired from other sources. The probability distribution over delay may depend on such factors as the operator's uncertainty about the best decision to make and the perceived criticality of a situation. We can extend the EVDI measure by considering uncertainty in delays, and by conditioning the probability distribution over length of delay on the quantity of evidence, and on such information as an operator's predicted uncertainty about the best action [Horvitz, 1995].

## 6 MODELING USER ACTIONS

In time-critical contexts, the goal of a display manager is to assist a user with taking the best action as soon as possible. As indicated by Equation 9, a key task in implementing approaches to display management based on EVDI is the development of a probabilistic model of an operator's beliefs and actions, as a function of displayed information. Assume that a gold-standard model indeed represents the preferences and probabilistic relationships of the best expertise available, but that the limited availability of experts and policies of an institution lead to situations that require nonexperts to make decisions. We wish to develop a display manager that continues to reason about how alternative quantities of information will change the nature and timeliness of such decisions.

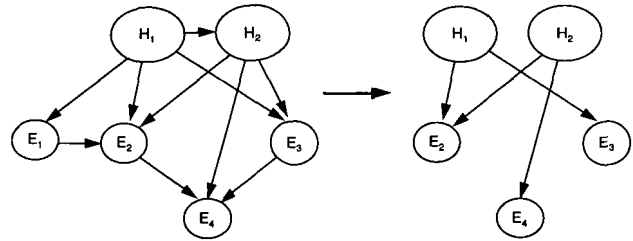

Figure 9: Constructing models of the operator's view of a system. Expert trainers of propulsion-systems operators found it useful to initiate the task of developing Bayesian models of a trainee's causal knowledge by pruning away subtle distinctions and dependencies from the gold-standard Bayesian networks and reassessing probabilities.

### 6.1 BAYESIAN MODELS OF USER BELIEFS

To control or design a display based on EVDI, we need to gain access to knowledge about $p(A_i|E, \xi)$ and $t(E)$ (or, more generally, $p(t_k|E, A, \xi)$). For simplification, let us assume that we have access to deterministic information about the amount of time required to review information as a function of the quantity of information displayed. We will focus on inference about user actions. We explored two approaches to modeling actions of operators as a function of their training and the data displayed.

One approach to modeling an operator's actions is to work with experts with experience with training people in their area of expertise, to construct directly models of user action. These models output probability distributions over user actions given assumed sets of observations displayed to an operator. Building probabilistic models that make inferences about $p(A|E, \xi)$ action is difficult; the modeling and assessment of actions typically requires the analysis of a large number of situations. It can be more efficient to construct Bayesian user models that represent, from an expert trainer's perspective, the causal probabilistic relationships assumed by trainees at varying levels of experience. These could be used to model the beliefs of users about anomalies given displayed data.

As part of the Vista effort to build models of user action for EVDI-based display, experts were asked to build Bayesian networks that could be used to make inferences about the beliefs of users in response to observations. We assessed from expert trainer's models of the causal knowledge and uncertain dependencies that represent expertise at different levels of training. The goal of this work was to compute the *probabilities that a user would assign* to the presence of system faults, after reviewing sets of observations. We found that expert trainers were comfortable building such probabilistic models representing users' beliefs about a monitored system. The models of user belief for Shuttle propulsion systems were tested and validating as performing like users at some level of training or experience.

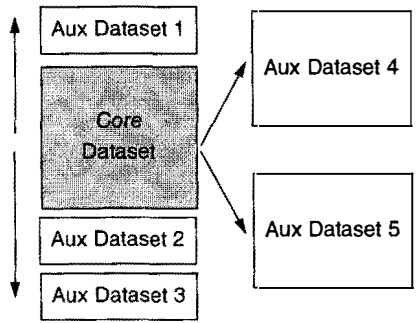

Figure 10: Decisions about auxiliary data. In one approach to display decision making, a decision-theoretic analysis is used to consider the benefits of displaying auxiliary information, beyond a standard core dataset.

As highlighted by Figure 9, experts found particularly useful a strategy of beginning the task of modeling a nonexpert's view of diagnosis and decision making by examining the gold-standard decision model and pruning away distinctions and dependencies considered to be absent in trainees' models of a monitored system. We use $p(H^D|E,\xi)$ to refer to the predictions of these Bayesian user models about the beliefs of a user given displayed information $E$.

### 6.2  FROM BELIEFS TO ACTIONS

Assume that we have a model that experts certify as providing accurate inference about an operator's beliefs as a function of evidence. How can we use $p(H^D|E,\xi)$ to compute the probability distribution over actions by the user, $p(A_i|E,\xi)$? If we had access to an operator's utility model, $u^D(A_i, H_j, t)$, representing the user's perception of outcomes and delays, and plausible set of actions and their dependencies, we could choose the action that maximizes the expected utility from the user's perspective and, thus, generate a "best" user action, $A^{D*}$. However, we cannot assume that users combine their beliefs in a coherent, decision-theoretic manner.

Several approaches show promise for allowing us to approximate $p(A_i|E,\xi)$. We assessed from experts user-model preferences to explore methods for combining user beliefs about faults into user actions. Experts identified differences between the preference model of a seasoned operator and a less-experienced person.

Given these models, we have explored several methods for mapping the inferred user's beliefs into the likelihood of actions, $p(A_i|E,\xi)$ for display management. In one approach, we assume that operators will attempt to maximize expected utility. In another, we assume more conservatively, that the likelihood of actions is a monotonically increasing function of the inferred expected utility of the actions given the user's preference model. Studies of the behavior of users with different levels of training will be useful for confirming and tuning such user models.

## 7  FLEXIBLE DISPLAY OPTIONS

EVRI and EVDI can be employed in a variety of ways to enhance the information displayed for time-critical monitoring. EVRI is more appropriate for use in situations where expert engineers will be making decisions, given the EVRI assumption of gold-standard decision making. We have found EVRI to be more appropriate for real-time monitoring of Shuttle propulsion systems because real-time decision making is limited to experts. EVDI promises to be useful for building effective systems for training and education. Applications of EVDI require greater amounts of engineering effort than EVRI because of the modeling of user's beliefs and actions required for EVDI.

The most general use of the display metrics involves a search through all combinations of information available for display. Rather than consider a search over all subsets of evidence, we can use the metrics at runtime or design time to make coarser display decisions, and employ approximate analyses based on single-step analysis or on a limited lookahead.

The metrics can be used as tools for evaluating and refining existing information layouts and display strategies. For example, we can examine the value of modifications to predefined clusters of related information, with an eye to optimizing a static display, or for doing real-time tailoring of the templates depending on context. We can also use the metrics to make decisions about the display of auxiliary clusters of information. Computer-based monitoring systems often display a default, core set of data, yet may have access to a large quantity of supportive information that is not typically relevant to decisions. System operators may wish to review a stable pattern of key variables, yet have access to supportive auxiliary information when it might change their decision. As an example, the main screen for Shuttle propulsion systems has not traditionally included information about sensor trend information, yet distinctions about trends are represented in the diagnostic models and can provide valuable information the likelihood of sensor failures. Decisions about auxiliary information have become especially salient when monitoring systems are upgraded from character-based displays to graphic workstations making available more display real-estate.

The EVRI and EVDI metrics can be used to determine when it is valuable to display auxiliary or more detailed sets of data, by considering the value versus the costs of displaying auxiliary information. For reasoning about the value of displaying auxiliary clusters of information, we continue to monitor telemetry and perform decision-theoretic inference. As highlighted in Figure 5, we compare the expected value of decisions with the core information display, $E^{core}$, versus with different extensions, $E^{core} + E_i^{aux}$. If the additional information leads to decisions with higher expected utility than the decision indicated by the information in the core display, the auxiliary information is displayed.

The display-management metrics can provide a formal foundation for such user-interface functionalities as the telescoping of templates of information about subsystems that was introduced in Vista-1. If we are monitoring several subsystems, as in the case of Shuttle propulsion systems, we can generate, for each subsystem, a set of templates of progressively greater detail and size. EVRI or EVDI can be used to make decisions about the escalation of templates, from a summary or core set of data to larger, more detailed expansions.

We do not have to model explicitly the costs of reviewing information to derive value from the metrics. For example, we can use EVRI or EVRI to search for minimal sets of evidence that are consistent with the action that has the greatest expected utility from the perspective of the gold-standard model. The minimal information strategy can be combined with means that allow user's to request with ease additional variables of interest. As another approach, we can employ EVRI to display auxiliary information whenever the additional information can change the best action taken, regardless of the criticality, or taking into consideration only gross indications of criticality.

In another application of the metrics, rather than using EVRI or EVDI to edit the displayed data, we seek to make more conservative decisions about the *highlighting* of information as a function of the situation and user. We again employ a myopic or limited lookahead search through all data being displayed, or evaluate classes of information as clusters. For the myopic analyses, we compute the value of revealing single pieces of data considered as hidden from the decision maker, in the context of the rest of the displayed data. Data can be highlighted with a range of intensity or color in accordance with the increasing value of data as indicated by the EVRI or EVDI. We can choose to highlight only the top $n$ observations, a number that may be determined dynamically by the quantity of data and time criticality. A similar analysis can be used to analyze the overall importance of different classes of clustered data to control the highlighting of different classes of data.

## 8    STATUS AND FUTURE WORK

We implemented in a prototype of the forthcoming Vista-III system the use of EVRI to highlight critical information with color to prioritize the review of data. We are currently exploring the use of EVRI for automated display of auxiliary information and for controlling the telescoping of information templates. We will be validating the new functionalities and expect the Vista-III system to be certified for flight at the Mission Control Center in the coming year. Overall, experts and trainees have been enthusiastic about the display management and decision-theoretic recommendations provided in the evolving Vista system.

We are working to further extend the display-management methods. In particular, we are exploring more sophisticated display-management techniques that take into consideration incompleteness or inaccuracy in the models of diagnosis and decision making used to compute advice. We wish to endow a reasoning system with explicit methods for evaluating the confidence in its diagnostic conclusions. Such self-awareness about model incompleteness, coupled with knowledge of when a human decision maker is likely to have deeper insights than the computer-based reasoner, will be valuable in building genuine decision-making associates.

We are concerned with the modeling effort that can be required to build systems based on EVDI. We are interested in developing more efficient methods for building and learning models that describe with fidelity the beliefs and actions of users in response to displayed information. We are excited about recent innovations in learning Bayesian networks from data and for combining expert models with data for building these models [Heckerman et al., 1994].

We look to future research in cognitive psychology for information about the costs associated with the review of information, including studies of how increased amounts of information and clutter can be expected to lead to delays, confusion, and, ultimately, to suboptimal decisions. Psychological studies can also give us insights about the value of such user-interface actions as highlighting displayed data with color, and costs associated with the instability of dynamically changing interface configurations. We are also interested in gaining deeper insights about the perceptions of users about automated display decisions versus completely manual access of auxiliary information. Such information about preferences and performance can allow us to build more sophisticated, effective display managers.

Beyond display management applications, EVRI and EVDI can be harnessed to make automated decisions about the most valuable information to transmit to remote decision makers. In such applications, we balance the value of information to enhance the timeliness or quality of decisions with the costs of transmitting the data. The methods provide a utility-based foundation for controlling the incremental transmission of time-critical information over limited-bandwidth channels.

We hope that the metrics and methods we described will be useful to others tackling issues surrounding real-time monitoring and control, and offline design of informational displays. We believe that the combination of automated decision-support and automated display systems will allow people to keep pace with the growing complexity and stakes associated with understanding and controlling the machines and software that we rely on to accomplish critical tasks.

## Acknowledgments

We appreciate the long-term contributions on the Vista Project by Jim Martin, Corinne Roukangas, Kevin Scott, Sampath Srinivas, and Cynthia Wells. We thank Jack Breese, Robert Fung, David Heckerman, Kathy Laskey, Paul Lehner, and Michael Shwe for useful feedback on Vista research. We are grateful to colleagues at the participating sites, including the NASA Johnson Space Center, Rockwell Palo Alto Laboratory, NASA Ames Research Center, and Stanford University. This work was supported by the NASA Johnson Space Center, Rockwell Science Center, the Rockwell Space Operations Company, and National Science Foundation Grant IRI-9108385.